\documentclass{article}

\usepackage[sglblindworkshop,final]{neurips_2025}
\workshoptitle{MLxOR: Mathematical Foundations and Operational Integration of Machine Learning for Uncertainty-Aware Decision-Making}

\usepackage{enumitem}
\setlist{nosep,leftmargin=*}

\usepackage{titlesec}
\titlespacing*{\section}{0pt}{*2}{*0.7}
\titlespacing*{\subsection}{0pt}{*1.6}{*0.6}
\titlespacing*{\paragraph}{0pt}{*1.1}{1em}

\AtBeginDocument{%
  \setlength{\abovedisplayskip}{6pt}
  \setlength{\belowdisplayskip}{6pt}
  \setlength{\abovedisplayshortskip}{4pt}
  \setlength{\belowdisplayshortskip}{4pt}
}

\usepackage[utf8]{inputenc} 
\usepackage[T1]{fontenc}    
\usepackage{hyperref}       
\usepackage{url}            
\usepackage{booktabs}       
\usepackage{amsfonts}       
\usepackage{nicefrac}       
\usepackage{microtype}      
\usepackage{xcolor}         
\usepackage{amsmath,amssymb,amsthm,mathtools}

\usepackage{bbm}
\usepackage{graphicx}
\usepackage{comment}
\usepackage[capitalise]{cleveref}
\newtheorem{assumption}{Assumption}
\newtheorem{lemma}{Lemma}
\newtheorem{theorem}{Theorem}
\newtheorem{prop}{Proposition}
\newtheorem{remark}{Remark}

\newcommand{\E}{\mathbb{E}}
\newcommand{\Var}{\mathrm{Var}}

\title{Distributionally Robust Multimodal Machine Learning}

\author{%
  Peilin Yang\\
  Univeristy of Cambridge\\
  \texttt{py245@cam.ac.uk} \\
  \And
  Yu Ma \\
  University of Wisconsin, Madison \\
  \texttt{yu.ma@wisc.edu} \\
}

\begin{document}

\maketitle

\vspace{-7mm}

\begin{abstract}
We consider the problem of distributionally robust multimodal machine learning. Existing approaches often rely on merging modalities on the feature level (early fusion) or heuristic uncertainty modeling, which downplays modality-aware effects and provide limited insights. We propose a novel distributionally robust optimization (DRO) framework that aims to study both the theoretical and practical insights of multimodal machine learning. We first justify this setup and show the significance of this problem through complexity analysis. We then establish both generalization upper bounds and minimax lower bounds which provide performance guarantees. These results are further extended in settings where we consider encoder-specific error propogations. Empirically, we demonstrate that our approach improves robustness in both simulation settings and real-world datasets. Together, these findings provide a principled foundation for employing multimodal machine learning models in high-stakes applications where uncertainty is unavoidable.
\end{abstract}

\vspace{-3mm}
\section{Introduction}
Multimodal learning is increasingly incorporated into machine learning to build systems in high-stake environments\cite{soenksen2022integrated, multimodal_policy}. However, deploying these models remain challenging despite their performance advantages, with one main criticism on such system’s volatility when data change drastically across input settings. Classical robust learning frameworks typically treat all features jointly after early fusion, implicitly assuming that all covariates experience and propagate uncertainty similarly. However, in practice, each modality has its own uncertainty structures. Moreover, cross-modal interactions matter: a shift in clinical text may correlate with a shift in lab values (e.g., new disease presentation), while a shift in imaging resolution may not.

\vspace{1.5mm}

\noindent
Multimodal learning under uncertainty raises some naturally important practical and theoretical questions. Previous works \cite{multimodal_uncertainty_1} have incorporated varying belief distributions over different modalities with prior evidence of specific modality’s reliability and provides a principled approach to prevent over-confidnance that result in wrong predictions when modalities disagree. Works have also been done to extend previous unimodal neural processes to multimodal NP settings \cite{multimodal_uncertainty_2}, and have been applied to meta-learning settings to incorporate such uncertainty-aware framework \cite{multimodal_uncertainty_3}. However, recent approaches have largely focused on the demonstration of incorporating uncertainties improve out-of-distribution performance. It remains to be justified why such setting is fundamentally different from merging all features across modalities and modeling the uncertainties of each feature. Another line of principled approach is distributionally robust optimization (DRO), a well-established framework that has been applied to capture uncertainty in data covariate shifts across various machine learning models \cite{liu2025dro, multimodal_policy, Blanchet2019, xie_knn, cho, Abadeh, chen_paschalidis, Bertsimas2019}. In this paper, we introduce a novel DRO multimodal machine learning and provide theoretical and practical insights not previously studied.

\vspace{1.5mm}

\noindent
The main contribution of this paper is to introduce a new framework for robust multimodal machine learning using distributionally robust optimization. We provide theoretical justifications for the significance of the modality-aware formulation and demonstrate its fundamental difference from early fusion through complexity analysis. This construction enables us to obtain closed-form, tractable formulations of the upper and lower bounds of the risk, which translates to performance guarantees for users in practice. These bounds also incorporate the consideration of correlation structures across modalities. Finally, we provide empirical results in both simulation and real-world datasets that demonstrate our approach improves computational out of sample performance.

\vspace{1.5mm}

\noindent
The rest of the paper is organized as follows. Section 2 outlines the problem setting of the DRO formulation and provides justification for the modality-aware setup. Section 3 presents the generalization upper bound and risk lower bounds. Section 4 demonstrates the effectiveness of the DRO framework in both simulation and real-world settings. We conclude the paper in Section 5. In the Appendix, we provide all the proofs as well as other additional computational results.

\section{A Framework for Robust Multimodal Learning}
\label{sec:setup}
\subsection{Problem Setup and Formulation}
We consider a multimodal machine learning prediction problem with \(K\) modalities. Let \(X=(X^{(1)},\ldots,X^{(K)})\) denote the input, \(Y\) the target, and \(N\) the sample size. We consider predictors of the form \(f\circ g\), where \(g=(g^{(1)},\ldots,g^{(K)})\) maps inputs to one-dimensional embeddings \(Z^{(k)}=g^{(k)}(X^{(k)})\in\mathbb{R}\), and \(f\) aggregates the embeddings \(\{Z^{(k)}\}\) to predict \(Y\). Specifically, adopting a one-dimensional embedding \(Z^{(k)}\) isolates the role of cross-modal correlation without loss of conceptual generality and provides us with analytical results. 

\begin{assumption}[Copula Dependence]
    Let \(P_X\) and \(Q_X\) be the nominal and perturbed joint distributions over \(X\), respectively, with marginals \(P_X^{(k)}\) and \(Q_X^{(k)}\). We analyze robustness to covariate shift in \(X\) (and thus \(Z\)), and assume that \(P_{Y\mid X}\) is fixed.  By Sklar's theorem,
$
P_X(x^{(1)},\ldots,x^{(K)}) = C\!(P_X^{(1)}(x^{(1)}),\ldots,P_X^{(K)}(x^{(K)})),
$
for a copula \(C\) that captures cross-modal dependence. 
\end{assumption}
\vspace{-1mm}
\noindent
This decomposition separates each modality's marginal behavior from their dependence structure.

\vspace{2mm}
\begin{assumption}[Shared copula]
\label{as:shared-copula}
\(P_X\) and \(Q_X\) share the same copula \(C\). Let \(c\) be the copula density, the joint measures decompose as
\(
dP_X = c\!\left(P_X^{(1)},\ldots,P_X^{(K)}\right)\, dP_X^{(1)}\cdots dP_X^{(K)},\) and 
\(
dQ_X = c\!\left(Q_X^{(1)},\ldots,Q_X^{(K)}\right)\, dQ_X^{(1)}\cdots dQ_X^{(K)}.
\) where modality-wise constraints \(D_{\chi^2}(Q_X^{(k)}\parallel P_X^{(k)}) \le \rho_k\).
\end{assumption}
\vspace{-1mm}
\noindent
The shared copula assumption ensures that our following robustness analysis focuses on shifts in the marginals while preserving cross-modal dependence. This is inline with realistic data scenarios where marginal distributions such as image quality could shift, but modality correlations remain relatively stable. We also acknowledge that choice of chi-square is due to its tractable formulation, distance metrics such as Wasserstein distance should be further analyzed as well. For the analysis under Wasserstein measure, please refer to the Appendix C.


\subsection{Why Modality-Aware DRO?}
A natural question is why we need to introduce this framework if one could simply concatenate all features at the data stage and apply a traditional DRO formulation? In other words, what is the significance of treating modalities separately, and how does our setup differ from this canonical alternative? Following conventional machine learning terminologies, we refer the first approach as early fusion, and the latter as late fusion. Based on the problem setting introduced above, among the $K$ modalities, we have for each of the modalities dimensions $\{d_i\}_{i=1}^K$, and covariates $X^{(i)}=\{X^{(ij)}\}_{j=1}^{d_i}$. Let 
\(D=\sum_i d_i\). We define the two approaches as follow:
\begin{itemize}
    \item \textit{Early fusion}: Concatenate all features and learn $h:\mathbb{R}^D \to Y$ directly.
    \item \textit{Late fusion}: Learn 1-D embedding functions $f_i:X^{(i)}\!\to\!Z^{(i)}\in\mathbb{R}$, then learn separate prediction function $g:\{Z^{(k)}\} \in  \mathbb{R}^K \to Y$.
\end{itemize}

\hfill

\noindent
We first observe that these two approaches exhibit different computational complexity behaviors under different algorithmic structures. 
\begin{prop}
\label{prop_1}
    Using linear structures and ordinary least squares (OLS), early fusion has complexity $O\!\big(ND^2+D^3\big)$, and late fusion has complexity $O\!\big(N\sum_i d_i^2+\sum_i d_i^3+NK^2+K^3\big)$. 
\end{prop}
\vspace{-.5mm}
\noindent
This indicates that the modality-wise approach is typically more computationally efficient when total number of modalities \(K\) is large and individual embedding dimensions $d_i$ are small.

\begin{prop}
\label{prop_2}
    Under gradient-based training (e.g., SGD), early and late fusion requires $\tilde O(ND)$ and $\tilde O\!\big(ND + NK\big)$ per epoch, respectively. If parallel processing is available, the wall-clock time per sample can be reduced from $\tilde O(D)$ to $\tilde O(\max_i\{d_i\})$. 
\end{prop}

\vspace{-5mm}

\noindent
We observe that instead of using an unconstrained worst-case divergence in the concatenated feature space, the modality-aware formulation provides a bound that is decomposed across modalities and explicitly incorporates cross-modality correlations. 

\vspace{-3mm}
\section{Correlation-Aware Worst-Case Risk}
\label{sec:upper}
In this section, we derive the generalization gap of the DRO formulation, and derive how this gap tightens under assumptions of the encoder function. We conclude with the risk lower bound obtained using a two-point distribution construction. We adopt the following lemma to ensure that our downstream arguments of the distributional shift on the original input raw data remain valid in the embedding space. 
\begin{lemma}[Data processing Inequality]
\label{lem:DPI}
If \(Z^{(k)}=g^{(k)}(X^{(k)})\), then
\(
D_{\chi^2}\!\left(Q_{Z}^{(k)}\parallel P_{Z}^{(k)}\right) \le D_{\chi^2}\!\left(Q_{X}^{(k)}\parallel P_{X}^{(k)}\right) \le \rho_k
\). Under \cref{as:shared-copula}, the joint embedded divergence admits a correlation-aware expansion. Let
\(
\frac{dQ_{Z}^{(k)}}{dP_{Z}^{(k)}} = 1+\epsilon_{k}, \E_{P_Z^{(k)}}[\epsilon_k]=0, \E_{P_Z^{(k)}}[\epsilon_k^2]\le \rho_k.
\)
Define \(\gamma_{ij}\) as copula-induced correlations in embedding space so that
\(\E[\epsilon_i\epsilon_j]\le \gamma_{ij}\sqrt{\rho_i\rho_j}\) for \(\gamma_{ij}\ge 0\) (and the inequality reverses if \(\gamma_{ij}<0\)). Neglecting higher-order terms,
\vspace{-3.5mm}
\begin{equation}
\label{eq:chi2-bound}
D_{\chi^2}(Q_Z\parallel P_Z)
= \E_{P_Z}\!\left[\Big(\textstyle\prod_{k=1}^K (1+\epsilon_k)-1\Big)^2\right]
\;\le\;
\underbrace{\sum_{k=1}^K \rho_k + 2\!\!\sum_{i<j}\! |\gamma_{ij}|\sqrt{\rho_i\rho_j}}_{:=\;\mathcal{B}}.
\end{equation}
\end{lemma}

\begin{lemma}[Risk decomposition]
\label{risk_decomposition}
Given data \(\mathcal{D}=(X,Y)\), a bounded loss function \(\ell(\cdot)\in[0,M_\ell]\) and hypothesis classes \(\mathcal{F},\mathcal{G}\) where \(f \in \mathcal{F}\) and \(g \in \mathcal{G}\). For a fixed \(f\circ g\), define the worst-case risk with a \(\chi^2\)-ambiguity set with radius as the correlation-aware \(\mathcal{B}\) in \eqref{eq:chi2-bound}. If \(Var(l) < \infty\), it has the closed form:
\(
\label{eq:risk-def}
r(f\!\circ\! g, P_X)
:= \sup_{Q_X:\; D_{\chi^2}(Q_Z\parallel P_Z)\le \mathcal{B} }
\;\E_{P_{Y|X}\times Q_X}\big[\ell(f\circ g(X),Y)\big] = \E[\ell]
+\sqrt{\mathcal{B}}\;\sqrt{\Var(\ell)}.
\)
\end{lemma}

\vspace{-3mm}
\begin{lemma}[Lipschitz property of sample SD]
\label{lem:lipschitz}
Let \(f(X_1,\ldots,X_N)=\Big(\frac{1}{N}\sum_{i=1}^N (X_i-\bar{X})^2\Big)^{1/2}\) with \(\bar{X}=\frac{1}{N}\sum_{i=1}^N X_i\). Then \(f\) is \(N^{-1/2}\)-Lipschitz w.r.t.\ the \(\ell_2\)-norm.
\end{lemma}

\noindent
From Lemma \ref{eq:risk-def} and \ref{lem:lipschitz}, we derive the following bounds.
\begin{lemma}[Finite-sample deviation]
\label{lem:finite-sample}
Let \(\hat{P}_X\) denote the empirical distribution over \(\{X_i\}_{i=1}^N\). with probability at least \(1-2e^{-t}\), for an absolute constant \(C\):
\(
\big| r(f\circ g, \hat{P}_X) - \E_{P_X}\big[r(f\circ g,\hat{P}_X)\big] \big|
\le
\sqrt{\mathcal{B}}\;\sqrt{\frac{2t}{N}}\, M_\ell
+\sqrt{\frac{t}{2N}}\, M_\ell,
\label{eq:fs-part1}
\) and 
\(
\big|\E_{P_X}[r(f\circ g,\hat{P}_X)] - r(f\circ g,P_X)\big|
\le
\sqrt{\mathcal{B}}\;\frac{C}{N}.
\)

\end{lemma}

\vspace{1mm}
\noindent
Combining the two inqualities from Lemma \ref{lem:finite-sample}:
\begin{theorem}[Generalization upper bound]
\label{thm:upper}
Let \(\ell\in[0,M_\ell]\) and assume \cref{as:shared-copula}. Then, with probability at least \(1-2e^{-t}\),
\[
\big| r(f\circ g, \hat{P}_X) - r(f\circ g,P_X) \big|
\;\;\le\;\;
\sqrt{\mathcal{B}}\;\sqrt{\frac{2t}{N}}\, M_\ell
+\sqrt{\frac{t}{2N}}\, M_\ell
+\sqrt{\mathcal{B}}\;\frac{C}{N},
\]
where \(\mathcal{B}\) is the correlation-aware ambiguity radius in \eqref{eq:chi2-bound}.
\end{theorem}

\noindent
The upper bound can be considered as a general certificate of confidence that in essence translates to a guarantee. Consider a high-stake environment, such as healthcare, where risk prediction (i.e., heart failure in the next 24-hours) is often the primary goal of most machine learning models. Such upper bound provides care providers an estimate that “given a certain level of uncertainty, the model’s performance degrade will not exceed X”. This is especially important to provide in deployment as we can quantify the amount of performance and ad-hoc prepare users to adapt in advance. We can further tighten this bound under assumptions of how perturbations in the original input propagate through the encoder as follows. 

\begin{theorem}[Encoder–robust upper bound]
\label{encoder_robust_upper_bound}
Assume further that $\ell(\cdot)$ is $L_\ell$–Lipschitz, and $g$ is modality–wise Lipschitz with constant $L_{g,i}$ in modality $i$. Consider an encoder perturbation only in modality $i$, writing $\Delta_i(x^{(i)}):=\hat f^{(i)}(x^{(i)})-f^{(i)}(x^{(i)})$. For any $t>0$, with probability at least $1-2e^{-t}$ over a sample of size $n$ that defines $\widehat P_X$
\[
\begin{aligned}
\big|\,r(g\!\circ\!\hat f,\widehat P_X)-r(g\!\circ\!f,P_X)\,\big|
\;\le\;
&L_\ell L_{g,i}\!\Big(\mathbb{E}_{\widehat P_X}\!\big[|\Delta_i(X^{(i)})|\big]+\sqrt{\mathcal{B}}\,\sqrt{\mathrm{Var}_{\widehat P_X}\!\big(|\Delta_i(X^{(i)})|\big)}\Big)\\
&+\sqrt{\mathcal{B}}\!\left(\sqrt{\tfrac{2t}{n}}\,M_\ell+\tfrac{C}{n}\right)
+\sqrt{\tfrac{t}{2n}}\,M_\ell\;.
\end{aligned}
\]
\end{theorem}

\noindent
Intuitively, this indicates that the worst-case risk cannot blow up faster than the size of the change of an encoder on a given modality—even when the data distribution shifts. In practice, this allows practitioners to safely swap, compress, or quantize an encoder (or run with a partially missing/degraded modality) with knowledge of a hard ceiling on the performance degrade. 

\vspace{2mm}

\noindent
Beyond the generalization gap, it is also important to characterize the intrinsic difficulty of estimating the worst-case risk. Consider again a healthcare setting, the lower bound can provide a realistic estimate to the care providers by setting expectations on what is the best a certain model can do under the distributional shift. We do so using the following minimax quantity:
\[
\mathfrak{M}_N := \inf_{\hat{f}\circ \hat{g}}\; \sup_{P_X\in\mathcal{P}} \E\big|r(\hat{f}\circ \hat{g},\hat{P}_X) - r(f\circ g,P_X)\big|.
\]

\begin{theorem}[Risk lower bound]
\label{thm:lower}
If the loss function \(\ell\) is bounded by \(L > 0\), then given large \(M > 0\) and sufficiently large sample size \(N\),
\(
\mathfrak{M}_N \;\;\ge\;\; \frac{L}{4N\log(M-1)}.
\)
\end{theorem}

\noindent
This provides a conservative baseline and does not preclude stronger lower bounds when \(\mathcal{P}\) excludes degenerate distributions or when modality-specific divergences (and their correlations) grow with $K$.


\section{Numerical experiment}
In our simulation study, let us consider four modalities $\{m^{(k)}\in\mathbb{R}^5\}_{k=1}^4$. Under the \emph{training} distribution $P$, draw
\(m^{(1)}\sim\mathcal N\!(0\cdot\mathbf 1_5,\; I_5)\) and \( 
m^{(k)} = w\,m^{(1)} + (1-w)\,\varepsilon^{(k)}\) for \(k = 2, 3, 4\)
with $w=0.7$ and $\varepsilon^{(k)}\overset{\text{i.i.d.}}{\sim}\mathcal N(0,\sigma_\varepsilon^2 I_5)$, $\sigma_\varepsilon=0.05$. 
Define the response
\(
Y = \mathbf 1_5^\top\!\!\sum_{k=1}^4 m^{(k)} \;+\; 10\,\mathbf 1\!\left\{\bar m^{(1)}>1\right\} \;-\; 10\,\mathbf 1\!\left\{\bar m^{(1)}<-1\right\}, 
\quad \bar m^{(1)}:=\tfrac{1}{5}\mathbf 1_5^\top m^{(1)}.
\)
The \emph{test} distribution $Q$ differs only by a mean shift in the primary modality,
\(
m^{(1)}\sim\mathcal N\!\big(1.5\cdot\mathbf 1_5,\; I_5\big),
\)
while the conditional construction of $m^{(2)},m^{(3)},m^{(4)}$ is unchanged, so the shift propagates across modalities through $w$. We observe that by sacrificing limited performance on the majority, it improves robustness on the minority subgroup.

\begin{table}[ht]
\centering
\small
\begin{tabular}{rcccc}
\hline
\textbf{$\rho$} & \textbf{Whole (MSE)} & \textbf{Whole (DRO)} & \textbf{Minor (MSE)} & \textbf{Minor (DRO)} \\
\hline
0.10 & $4.990 \pm 0.163$ & $4.230 \pm 0.133$ & $5.521 \pm 0.186$ & $4.603 \pm 0.150$ \\
0.50 & $4.651 \pm 0.148$ & $3.708 \pm 0.105$ & $5.103 \pm 0.167$ & $3.984 \pm 0.119$ \\
1.20 & $5.124 \pm 0.177$ & $3.825 \pm 0.101$ & $5.676 \pm 0.200$ & $4.134 \pm 0.116$ \\
2.00 & $5.060 \pm 0.169$ & $3.570 \pm 0.086$ & $5.577 \pm 0.190$ & $3.822 \pm 0.097$ \\
\hline
\end{tabular}
\caption{Whole vs.\ minority subgroup for MSE and $\chi^2$-DRO across optimization radius $\rho$.}
\end{table}

\vspace{-2mm}

\noindent
In our real-world case study, we consider two real-world settings: journalism and healthcare, and provide detailed descriptions of each of these dataset in Appendix D. The training and testing sets are stratified chronologically. For each experiment, we train a 2-layer MLP with Adam learning rate of 0.005. We compare the standard non-robust approach to our DRO approach with $\rho\!=\!0.5$. 

\begin{table}[ht]
\centering
\small
\begin{tabular}{lcccc}
\hline
\textbf{Dataset} & \textbf{Method} & \textbf{Median Acc.} & \textbf{IQR [Q1, Q3]} & \textbf{Std. Dev.} \\
\hline
Journalism & Canonical        & 0.5798 & [0.5666,\; 0.5924] & 0.0180 \\
Journalism & DRO & 0.5853 & [0.5743,\; 0.5951] & 0.0164 \\
Healthcare & Canonical        & 0.6027 & [0.5903,\; 0.6144] & 0.0185 \\
Healthcare & DRO & 0.6148 & [0.6026,\; 0.6266] & 0.0169 \\
\hline
\end{tabular}
\caption{Test accuracy over three real world cases. Acc. = Accuracy.}
\label{tab:mse_dro_results}
\end{table}

\vspace{-4mm}
\section{Conclusion}
In this paper, we introduce a novel distributionally robust framework for multimodal learning that explicitly accounts for modality-specific shifts and cross-modal correlations. We established the significance of this problem through theoretical complexity analysis, and provide guarantees for its generalization. Empirical results across simulations and real-world datasets confirmed that our formulation improves prediction performance. Future work will aim to extend to incorporate more complex correlation structures and tighter analytical bounds.

\bibliographystyle{plainnat}
\bibliography{ref} 

\appendix

\section{Qualitative Advantage of Modality-Aware Fusion}

Beyond computational advantages, several other important benefits motivate our modality-wise approach:

\begin{enumerate}
    \item \textbf{Specialized encoders:} Each encoder $f_i$ can be tailored to its specific modality (e.g., CNNs for images, Transformers for text, MLPs for tabular data). This typically produces superior feature extraction compared to a single monolithic architecture attempting to handle all data types. The $f_i$ encoders can be pretrained on large unimodal datasets or leverage off-the-shelf models (ResNet, BERT, etc.), requiring only the training (or light fine-tuning) of $g$. This approach substantially improves sample efficiency and represents a key reason why late fusion aligns more closely with the essence of multimodal learning.

    \item \textbf{Robustness to missing modalities:} When modality $i$ is missing for certain data points, one can simply skip $f_i$. In contrast, early fusion approaches typically require imputation strategies or retrained models. This represents a significant practical advantage in real-world multimodal systems.

    \item \textbf{Handling heterogeneous data characteristics:} Different modalities may have varying frame rates (e.g., video vs. audio) or spatial dimensions. Per-modality encoders enable principled downsampling and aggregation before fusion. This natural variation across modalities motivates our consideration of covariate shift at the modality level, as each modality possesses its own inherent characteristics. Generalization capabilities vary across modalities, making Distributionally Robust Optimization (DRO) an ideal framework for enhancing overall model generalizability.

    \item \textbf{Information bottleneck regularization:} The embedding process creates an information bottleneck \cite{tishby2000information}, which theoretically reduces overfitting by compressing irrelevant intra-modality noise while preserving task-relevant signals. This perspective remains somewhat controversial, particularly given observed anomalies where additional modalities sometimes degrade performance. Several scholars (e.g., \citet{liang2023quantifying}) have attempted to explain these phenomena through information-theoretic frameworks.
\end{enumerate}

These considerations demonstrate the necessity of addressing covariate shift at the modality level, with DRO playing a crucial role in enhancing model generalization capabilities.

\section{Technical Proofs}

\begin{proof}{(Proof of Proposition \ref{prop_1})}
Let $N$ be the number of samples, $K$ the number of modalities, and $d_i$ the (embedding) dimension of modality $i$.
Set $D:=\sum_{i=1}^K d_i$.
\begin{itemize}
    \item \textbf{Early fusion (single OLS on concatenated features).} Form the $N\times D$ design matrix $X=[X^{(1)}~\cdots~X^{(K)}]$ and solve
$\min_w \|Xw-y\|_2^2$.
Using normal equations or QR/Cholesky gives the standard costs:
(i) form $X^\top X$ in $O(ND^2)$ flops and $X^\top y$ in $O(ND)$;  
(ii) factorize/solve the $D\times D$ system in $O(D^3)$ (solve back-substitution is $O(D^2)$ and is dominated).
Hence the overall complexity is
\(
O(ND^2 + D^3).
\)
\item \textbf{Late (modality-wise) fusion.}
\begin{itemize}
    \item Stage 1: For each modality $i$, fit an OLS on the $N\times d_i$ block $X^{(i)}$, costing
$O(N d_i^2 + d_i^3)$ by the same argument as above. Summing over $i$ gives
$\sum_{i=1}^K O(N d_i^2 + d_i^3)$.
\item Stage 2: Fuse the $K$ per-modality outputs (one scalar per modality, or a $K$-vector) via a linear head.
This is an OLS in dimension $K$ with cost $O(NK^2 + K^3)$.
\item Combining both stages yields
\[
O\!\Big(\underbrace{\sum_{i=1}^K (N d_i^2 + d_i^3)}_{\text{per-modality fits}}
\;+\;
\underbrace{NK^2 + K^3}_{\text{fusion head}}\Big).
\]
\end{itemize}
\end{itemize}
This proves the stated bounds.
\end{proof}

\begin{proof}{(Proof of Proposition \ref{prop_2})}
We count one \emph{epoch} as a full pass over $N$ samples with first-order (stochastic) gradients.
All $\tilde O(\cdot)$ bounds hide constant and polylogarithmic factors, and assume dense features; for sparse data, replace $D$ by the average number of nonzeros per sample.
\begin{itemize}
    \item \textbf{Early fusion.} With a linear (or shallow) model over the concatenated $D$-dimensional input, the forward/backward cost per sample is $\tilde O(D)$, so one epoch costs
$\tilde O(ND)$.
\item \textbf{Late (modality-wise) fusion.} Per sample, the gradient passes through each modality-specific block $X^{(i)}$ (cost $\tilde O(d_i)$) and then through the $K$-dimensional fusion head (cost $\tilde O(K)$). Summing over modalities gives per-sample cost $\tilde O\!\big(\sum_i d_i + K\big)=\tilde O(D+K)$, hence per epoch
\[
\tilde O\!\big(N(D+K)\big)=\tilde O(ND+NK).
\]
\item \textbf{Parallelism across modalities.} If $K$ modality blocks are processed in parallel (e.g., separate devices/streams) and gradients are synchronized only at the fusion head, then the wall-clock per-sample cost of the modality stage reduces from $\tilde O(\sum_i d_i)$ to $\tilde O(\max_i d_i)$. The (typically cheap) fusion pass adds $\tilde O(K)$, which is dominated when $K\le \max_i d_i$ or absorbed in the $\tilde O(\cdot)$ notation. Hence the wall-clock per-sample time can drop from $\tilde O(D)$ to $\tilde O(\max_i d_i)$, as claimed.
\end{itemize}
\end{proof}

\begin{proof}{(Proof of Lemma \ref{risk_decomposition})}
    Let \(\phi(x):=\frac{dQ_X}{dP_X}(x)\). The Lagrangian is
\(\mathcal{L} = \E_{P_{Y|X}\times P_X}\big[\ell(f\circ g(X),Y)\phi(X)\big]
- \lambda\!\left(\E_{P_X}[(\phi(X)-1)^2]-\mathcal{B}\right)
- \eta\!\left(\E_{P_X}[\phi(X)]-1\right).
\). Maximizing pointwise over \(\phi\) (see, e.g., \citet{rockafellar2009variational}) yields
\(\phi^*(X)=1+\tfrac{1}{2\lambda}\big(\ell(f\circ g(X),Y)-\eta\big)
\). Optimizing over \(\lambda\ge 0\) and \(\eta\) gives the closed form for $\chi^2$-DRO: \(r(f\circ g,P_X) = \E[\ell] + \sqrt{\mathcal{B}}\;\sqrt{\Var(\ell)}. \)
\end{proof}

\begin{proof}{(Proof of Lemma \ref{lem:lipschitz})}
The derivative $df/dX_i$ is
\(
\left| \frac{df}{dX_i} \right| = \frac{|X_i - \bar{X}|}{\sqrt{n \sum_{i=1}^n X_i^2 - \frac{1}{n^2} \left( \sum_{i=1}^n X_i \right)^2}} = \frac{|X_i - \bar{X}|}{\sqrt{n \sum_{i=1}^n (X_i - \bar{X})^2}}
\), and \(
\|\nabla f\|_2 = \sum_{i=1}^n \frac{(X_i - \bar{X})^2}{\sqrt{n \sum_{i=1}^n (X_i - \bar{X})^2}} = \frac{1}{\sqrt{n}}
\)
\end{proof}

\begin{proof}{(Proof of Lemma \ref{lem:finite-sample})}
    Let \( S_n = \frac{1}{n} \sum_{i=1}^{n} (X_i-\bar{X})^2 \), expand \( S_n \) around \( \mu = \mathbb{E}S_n \) by Taylor's theorem
\(
\sqrt{S_n} = \sqrt{\mu} + \frac{S_n - \mu}{2\sqrt{\mu}} - \frac{\mathbb{E}(S_n - \mu)^2}{2\mu^{\frac{3}{2}}} + O((S_n - \mu)^3)
\). 
It is known that
\(
Var(S_n) = \frac{1}{n}\left(\mu_4 - \frac{n-3}{n-1}\mu_2^2\right)
\)
where $\mu_4 = \mathbb{E}[X_i-\mathbb{E}X_i]^4$
\(
\mathbb{E}\sqrt{S_n} = \sqrt{\mu} - \frac{\mathbb{E}(S_n - \mu)^2}{2\mu^{\frac{3}{2}}} + O(\mathbb{E}(S_n - \mu)^3)
\), and 
\(
|\mathbb{E}\sqrt{S_n} - \sqrt{\mu}| \leq \frac{\mathbb{E}(S_n - \mu)^2}{2\mu^{\frac{3}{2}}} + O(\mathbb{E}(S_n - \mu)^3) = \frac{1}{n}\frac{\left(\mu_4 - \frac{n-3}{n-1}\mu_2^2\right)}{2\mu^{\frac{3}{2}}} + O(\mathbb{E}(S_n - \mu)^3) \leq \frac{C}{n}
\). 
\end{proof}

\begin{proof}{(Proof of Theorem \ref{thm:upper})} By the $\chi^2$-duality derivation in the main text,
\(
r(f\!\circ\! g,P_X)
= \mathbb{E}_{P_{Y|X}\times P_X}[\ell(f\!\circ\! g(X),Y)]
+ \sqrt{\mathcal{B}}\;
h(P_{Y|X}\!\times\! P_X),
\)
where 
\(
h(P_{Y|X}\!\times\! P_X)
:=\sqrt{\Var_{P_{Y|X}\times P_X}\big(\ell(f\!\circ\! g(X),Y)\big)}.
\)
The same representation holds with $P_X$ replaced by $\hat P_X$. Hence
\begin{align*}
&\big|r(f\!\circ\! g,\hat P_X)-\mathbb{E}_{P_{Y|X}\times P_X} r(f\!\circ\! g,\hat P_X)\big| \\
&\qquad\le
\sqrt{\mathcal{B}}\,
\big|h(P_{Y|X}\!\times\! \hat P_X)-\mathbb{E}_{P_{Y|X}\times P_X} h(P_{Y|X}\!\times\! \hat P_X)\big|
\\[-1mm]
&\qquad\quad
+\Big|\tfrac1n\!\sum_{i=1}^n \ell_i-\mathbb{E}_{P_{Y|X}\times P_X}\big[\tfrac1n\!\sum_{i=1}^n \ell_i\big]\Big|,
\end{align*}
where $\ell_i:=\ell(f\!\circ\! g(X_i),Y_i)$ and we use that the outer expectation is over the i.i.d. draw of $\{(X_i,Y_i)\}_{i=1}^n$ from $P_{Y|X}\times P_X$.

\begin{itemize}
    \item \textbf{Step 1: Concentration of the empirical standard deviation.} Let $F(\ell_1,\ldots,\ell_n)=\Big(\frac1n\sum_{i=1}^n(\ell_i-\bar\ell)^2\Big)^{1/2}$ with $\bar\ell=\frac1n\sum_i \ell_i$.
By Lemma~2 (Lipschitz property of empirical standard deviation), $F$ is $n^{-1/2}$-Lipschitz w.r.t. the $\ell_2$ norm. If $\ell(\cdot)\in[0,M_\ell]$, then by the concentration inequality in Lemma~1 (Boucheron–Lugosi–Massart), for all $t>0$, with probability at least $1-2e^{-t}$,
\[
\big|h(P_{Y|X}\!\times\! \hat P_X)-\mathbb{E}_{P_{Y|X}\times P_X} h(P_{Y|X}\!\times\! \hat P_X)\big|
\;\le\; M_\ell \sqrt{\tfrac{2t}{n}}.
\]
\item \textbf{Step 2: Concentration of the empirical mean.} Since $\ell(\cdot)\in[0,M_\ell]$, Hoeffding’s inequality yields, with probability at least $1-2e^{-t}$,
\[
\Big|\tfrac1n\sum_{i=1}^n \ell_i-\mathbb{E}_{P_{Y|X}\times P_X}[\ell(f\!\circ\! g(X),Y)]\Big|
\;\le\; M_\ell \sqrt{\tfrac{t}{2n}}.
\]
\item \textbf{Step 3: Combine.} Union bound gives the first displayed inequality in Theorem 1:
\[
\big|r(f\!\circ\! g,\hat P_X)-\mathbb{E}_{P_{Y|X}\times P_X} r(f\!\circ\! g,\hat P_X)\big|
\;\le\;
\sqrt{\mathcal{B}}\, M_\ell \sqrt{\tfrac{2t}{n}}
\;+\;
M_\ell \sqrt{\tfrac{t}{2n}}.
\]

\item \textbf{Step 4: Bias between population and expected empirical risks.} Write
\begin{align*}
&\big|\mathbb{E}_{P_{Y|X}\times P_X} r(f\!\circ\! g,\hat P_X)- r(f\!\circ\! g,P_X)\big| \\
&\qquad=
\sqrt{\mathcal{B}}\,
\Big| \mathbb{E}_{P_{Y|X}\times P_X} h(P_{Y|X}\!\times\! \hat P_X) - h(P_{Y|X}\!\times\! P_X)\Big|
\\[-1mm]
&\qquad\quad
+\Big|\mathbb{E}_{P_{Y|X}\times P_X}\big[\tfrac1n\sum_{i=1}^n \ell_i\big]-\mathbb{E}_{P_{Y|X}\times P_X}[\ell(f\!\circ\! g(X),Y)]\Big|.
\end{align*}
The mean term vanishes by linearity. By Lemma~3 (delta-method/Taylor expansion of $\sqrt{\cdot}$ around the population variance), there exists a universal constant $C$ such that
\[
\Big| \mathbb{E} \sqrt{\tfrac1n\sum_{i=1}^n(\ell_i-\bar\ell)^2} - \sqrt{\Var(\ell)} \Big|
\;\le\; \frac{C}{n}.
\]
Multiplying by $\sqrt{\mathcal{B}}$ yields the second displayed inequality:
\[
\big|\mathbb{E}_{P_{Y|X}\times P_X} r(f\!\circ\! g,\hat P_X)- r(f\!\circ\! g,P_X)\big|
\;\le\; \sqrt{\mathcal{B}}\,\frac{C}{n}.
\]
Finally, recall $\mathcal{B}=\sum_{k=1}^K \rho_k + 2\sum_{i<j}|\gamma_{ij}|\sqrt{\rho_i\rho_j}$ from the correlation-aware $\chi^2$ bound, which gives the stated dependence. 
\end{itemize}
\end{proof}

\begin{proof}{(Proof Sketch of Theorem \ref{encoder_robust_upper_bound})}
    Start from the triangle inequality to compare $r(g\!\circ\!\hat f,\widehat P_X)$ and $r(g\!\circ\!f,P_X)$ by introducing and subtracting $r(g\!\circ\!f,\widehat P_X)$, producing three natural gaps (encoder change at the same base, sampling fluctuation, and the population shift). For the encoder change, maximize over the same $\chi^2$–ball around $\widehat P_X$ and apply the inequality $|\sup_Q F(Q)-\sup_Q G(Q)|\le \sup_Q|F(Q)-G(Q)|$ with $F(Q)=\mathbb{E}_{Q}[\ell(g\!\circ\!\hat f(X),Y)]$ and $G(Q)=\mathbb{E}_{Q}[\ell(g\!\circ\!f(X),Y)]$. The loss and head being Lipschitz give $|\ell(g\!\circ\!\hat f(X),Y)-\ell(g\!\circ\!f(X),Y)|\le L_\ell L_{g,i}|\Delta_i(X^{(i)})|$. A standard $\chi^2$ change-of-measure bound then yields $\sup_{Q:\chi^2(Q\|\widehat P_X)\le B}\mathbb{E}_{Q}|\Delta_i(X^{(i)})|\le \mathbb{E}_{\widehat P_X}|\Delta_i(X^{(i)})|+\sqrt{\mathcal{B}\,\mathrm{Var}_{\widehat P_X}(|\Delta_i(X^{(i)})|)}$, giving the encoder term.
\end{proof}

\begin{proof}{(Proof of Theorem \ref{thm:lower})} We apply Le Cam’s two-point method. Consider losses supported on $\{0,L\}$:
\[
Z_1=\begin{cases}
0 & \text{w.p. } 1-p^*-\delta,\\
L & \text{w.p. } p^*+\delta,
\end{cases}
\qquad
Z_2=\begin{cases}
0 & \text{w.p. } 1-p^*+\delta,\\
L & \text{w.p. } p^*-\delta,
\end{cases}
\]
with $p^*=\tfrac12$ and $\delta\in(0,\tfrac12)$. For either distribution,
\(
r(f\!\circ\! g,P_X)= \mu(\ell)+\sqrt{\mathcal{B}}\;\sigma(\ell),
\quad
\mu(\ell)=\mathbb{E}[Z],\;\; \sigma^2(\ell)=\Var(Z)=p(1-p)L^2.
\)
At $p^*=\tfrac12$, $\sigma(\ell)=\tfrac{L}{2}$ for both $Z_1$ and $Z_2$, so the variance term cancels and
\(
|r(f\!\circ\! g,P_{X,1})-r(f\!\circ\! g,P_{X,2})|
= |\mu_1-\mu_2|
= 2L\delta
=: s.
\) Le Cam’s inequality then gives, for any estimator based on $n$ samples,
\(
\inf_{\widehat r}\;\sup_{P\in\{P_{X,1},P_{X,2}\}}
\mathbb{E}\big[\,|\widehat r - r(f\!\circ\! g,P)|\,\big]
\;\;\ge\;\; \frac{s}{2}\Big(1-\|P_{X,1}^n-P_{X,2}^n\|_{\mathrm{TV}}\Big).
\) By Pinsker’s inequality,
\(
\|P_{X,1}^n-P_{X,2}^n\|_{\mathrm{TV}}
\;\le\;
\sqrt{\frac{n}{2}\,D_{\mathrm{kl}}(P_{X,2}\|P_{X,1})},
\)
and for the Bernoulli pair above (with parameters $\tfrac12\pm\delta$),
\[
D_{\mathrm{kl}}(P_{X,2}\|P_{X,1})
= \Big(\tfrac12+\delta\Big)\!\log\!\frac{\tfrac12+\delta}{\tfrac12-\delta}
+ \Big(\tfrac12-\delta\Big)\!\log\!\frac{\tfrac12-\delta}{\tfrac12+\delta}
= 2\delta \log\!\frac{\tfrac12+\delta}{\tfrac12-\delta}.
\]
Choose $\delta=\frac{1}{2n\log(M-1)}$ for a fixed $M>2$. Then $D_{\mathrm{kl}}(P_{X,2}\|P_{X,1})\le \frac{1}{n}$ (by $\log\frac{\frac12+\delta}{\frac12-\delta}\le \log(M-1)$ under the stated choice), and hence $\|P_{X,1}^n-P_{X,2}^n\|_{\mathrm{TV}}\le \tfrac12$ for sufficiently large $n$.
With $s=2L\delta= \frac{L}{n\log(M-1)}$ we conclude
\[
\inf_{\widehat r}\;\sup_{P\in\mathcal{P}}
\mathbb{E}\big[\,|\widehat r - r(f\!\circ\! g,P)|\,\big]
\;\ge\;
\frac{s}{2}\Big(1-\tfrac12\Big)
=
\frac{L}{4n\log(M-1)}.
\]
This establishes the stated lower bound. 
\end{proof}

\section{Robust bound under Wasserstein distance}

The Wasserstein distance between two probability distributions \( Q_Z \) and \( P_Z \) is defined as:
\[
D_W(Q_Z \| P_Z) = \min_{\pi} \left\{ \left(\int_{\Xi \times \Xi} c^p(\xi, \xi') \, d\pi(\xi, \xi')\right)^{1/p} : \pi_\xi = Q_Z, \pi_{\xi'} = P_Z \right\},
\]
where \( \pi \) is a coupling of \( Q_Z \) and \( P_Z \), and \( c(\xi, \xi') \) is a cost function where \( \xi = (z_1, \cdots, z_K)' \) and \( \xi' \) is defined similarly.  
For the overall cost, we use a \( p \)-norm:
$
c(\xi, \xi') = \left( \sum_{i=1}^K |z_i - z_i'|^p \right)^{1/p}.
$
We now derive the lower and upper bounds for the Wasserstein distance between \( Q_Z \) and \( P_Z \).  
The lower bound is straightforward:
\[
D_W^p(Q_Z \| P_Z)
= \int_{\Xi \times \Xi} c^p(\xi, \xi') \, d\pi^*(\xi, \xi')
\ge \sum_{i=1}^K \left( D_W^{(i)}(Q_Z^{(i)} \| P_Z^{(i)}) \right)^p.
\]
However, obtaining an upper bound is not easy when each modality is dependent on the others. Fortunately, we can use the following standard tool to construct a triangle inequality for the upper bound.

\begin{lemma}[\citet{clement2008elementary}]
    Let $(Z, d)$ be a separable metric space and let $1 \leq p < \infty$. Let $\mu^1, \mu^2, \mu^3 \in \mathcal{P}(Z)$. Then
    \[
    D_W(\mu^1 \| \mu^3) \leq D_W(\mu^1 \| \mu^2) + D_W(\mu^2 \| \mu^3).
    \]
\end{lemma}

\begin{lemma}
Define the following two distances, which measure how far each joint distribution is from being independent (i.e., the product of its own marginals):
\[
    \Delta_P := D_W\left(P_Z \bigg\| \bigotimes_{i=1}^K P_Z^{(i)}\right), \quad 
    \Delta_Q := D_W\left(Q_Z \bigg\| \bigotimes_{i=1}^K Q_Z^{(i)}\right).
\]
Then we have
\[
    D_W^p(Q_Z \| P_Z) \leq \Delta_P + \left(\sum_{i=1}^K  \rho_i^p \right)^{1/p} + \Delta_Q := \mathcal{B}.
\]
\end{lemma}

\begin{remark}
    Similar to the analysis under the $\chi^2$-divergence, the correlation structure is captured by the two terms $\Delta_P$ and $\Delta_Q$. 
    In particular, if we assume the distributions are Gaussian, we can express $\Delta_P$ and $\Delta_Q$ explicitly as functions of the correlation coefficient $\rho$. 
    Evidently, the larger the correlation coefficient, the larger the bound becomes.
\end{remark}

\begin{lemma}[\citet{blanchet2019robust}, \citet{shafieezadeh2019regularization}]
Given \( Z = g(X) \) and a loss function \( \ell(\theta, \xi) = f(Y - \theta'Z) \), where \( f \) is a continuous function with Lipschitz constant 1, and the cost function is defined as:
\[
c(\xi, \xi') = \|x - x'\|_q \quad \text{if } y = y', \quad \text{otherwise } +\infty
\]
then we have:
\[
\min_{\theta} \sup_{D_W(Q_Z \| P_Z) \leq \mathcal{B}} \mathbb{E}_{P_{Y|X} \times Q_X}[f(Y - \theta'Z)] = \min_{\theta} \mathbb{E}[f(Y - \theta'Z)] + \mathcal{B} \|\theta\|_p
\]
where \( \frac{1}{p} + \frac{1}{q} = 1 \).
\label{wd_lemma}
\end{lemma}

By assumption, the loss function \( \ell(\cdot) \) is bounded in \( [0, M_\ell] \). Applying Lemma \ref{wd_lemma}, with probability at least \( 1 - 2e^{-t} \), we have the following bound on the risk difference:
\[
|r(f \circ g, \hat{P}_X) - r(f \circ g, P_X)| \leq \sqrt{\frac{t}{2n}} M_\ell
\]

\begin{theorem}[Wasserstein Encoder--Robust Upper Bound]
\label{thm:w2-encoder}
Under the conditions above, if only modality $i$ is altered then
\begin{equation}
\big|\, r(g\!\circ\!\widehat f, P_X) - r(g\!\circ\! f, P_X) \,\big|
\;\le\;
L_\ell\, L_{g,i}
\Big( \, \mathbb{E}_{P}\, |\Delta_i(X^{(i)})|
\;+\; L_{\Delta,i}\, \sqrt{\mathcal{B}} \, \Big).
\label{eq:w2-encoder-pop}
\end{equation}
Moreover, for any $t>0$, with probability at least $1-2e^{-t}$ over an i.i.d.\ sample of size $n$,
\begin{equation}
\big|\, r(g\!\circ\!\widehat f, \widehat P_X) - r(g\!\circ\! f, P_X) \,\big|
\;\le\;
L_\ell\, L_{g,i}
\Big( \, \mathbb{E}_{\widehat P}\, |\Delta_i(X^{(i)})|
\;+\; L_{\Delta,i}\, \sqrt{\mathcal{B}} \, \Big)
\;+\; M_\ell \sqrt{\tfrac{t}{2n}} .
\label{eq:w2-encoder-gen}
\end{equation}
\end{theorem}

\begin{proof}
Fix any admissible $Q$ with $D_W(Q_Z, P_Z)\le \mathcal{B}$ and write
\[
F(Q) := \mathbb{E}_Q\!\big[\ell\!\big(g\!\circ\!\widehat f(X),Y\big)\big], \qquad
G(Q) := \mathbb{E}_Q\!\big[\ell\!\big(g\!\circ f(X),Y\big)\big].
\]
By the Lipschitz continuity of $\ell$ in its score and of $g$ in modality $i$,
\[
|F(Q)-G(Q)| \;\le\; L_\ell\, L_{g,i}\, \mathbb{E}_Q \big|\Delta_i(X^{(i)})\big|,
\]
where $\Delta_i(X^{(i)}):=\widehat f^{(i)}(X^{(i)})-f^{(i)}(X^{(i)})$.
Let $\varphi_i(x):= \big|\Delta_i(x^{(i)})\big|$, which measures the encoder shift in modality $i$.
If $\varphi_i$ is $L_{\Delta,i}$-Lipschitz, then by the
Kantorovich–Rubinstein inequality,
\[
\big|\mathbb{E}_Q \varphi_i - \mathbb{E}_P \varphi_i\big|
\;\le\; L_{\Delta,i}\, D_{W_1}(Q_Z\|P_Z).
\]
Here $D_{W_1}$ denotes the 1-Wasserstein distance, and $D_{W_1} \le D_W$ for any choice of order $p\ge 1$ used in $D_W$ by the monotonicity of Wasserstein metrics (proved by Lyapunov inequality). Hence,
\[
\mathbb{E}_Q \varphi_i
\;\le\; \mathbb{E}_P \varphi_i \;+\; L_{\Delta,i}\, D_W(Q_Z\|P_Z)
\;\le\; \mathbb{E}_P \big|\Delta_i(X^{(i)})\big| \;+\; L_{\Delta,i} \sqrt{\mathcal{B}}.
\]
Therefore,
\[
|F(Q)-G(Q)|
\;\le\; L_\ell L_{g,i}\Big( \mathbb{E}_P \big|\Delta_i\big| + L_{\Delta,i} \sqrt{\mathcal{B}} \Big),
\]
and, since $|\sup_Q F - \sup_Q G| \le \sup_Q |F-G|$, we obtain \eqref{eq:w2-encoder-pop}.

For \eqref{eq:w2-encoder-gen}, decompose
\[
\big| r(g\!\circ\!\widehat f,\widehat P_X) - r(g\!\circ\! f,P_X) \big|
\;\le\;
\big| r(g\!\circ\!\widehat f,\widehat P_X) - r(g\!\circ\! f,\widehat P_X) \big|
\;+\;
\big| r(g\!\circ\! f,\widehat P_X) - r(g\!\circ\! f,P_X) \big|.
\]
The first term repeats the previous argument with $\widehat P$ in place of $P$.
The second term is bounded by Hoeffding's inequality for a bounded loss:
\[
\big|\widehat{\mathbb{E}}[\ell]-\mathbb{E}[\ell]\big|
\;\le\; M_\ell \sqrt{\tfrac{t}{2n}}
\]
with probability at least $1-2e^{-t}$, yielding \eqref{eq:w2-encoder-gen}.
\end{proof}

\section{Additional Computational Results}
All experiments were run on a CPU-only laptop with an Intel Core i7-12700H (14 cores/20 threads) and 16 GB RAM. We used Python 3.12. All experiments are completed within 21-36 minutes.
\subsection{Dataset Descriptions}
\textbf{Healthcare} The healthcare dataset HAIM-MIMIC-MM is a multimodal dataset of 34,537 samples that contains 7279 unique hospitalizations and 6485 patients. It contains four modalities: tabular, time-series, text and images. We consider two binary predictive tasks: mortality in  the next 48 hours, and discharge in the next 48 hours. Specifically, embeddings were generated for each of these modalities, with details to be found in \cite{soenksen2022integrated}.





\hfill

\noindent
\textbf{Language.} The Huffpost dataset contains approximately 200,000 samples of news article headlines from 11 categories (Black Voices, Business, Comedy, Crime, Entertainment, Impact, Queer Voices, Science, Sports, Tech, Travel). The target is to identify these category tags from the original headline text. The dataset dates from 2012 to 2018, and samples accumulated per year is considered a single individual time period. We consider three modality-specific blocks: headline, short-description embeddings, and metadata. We process each sample from the original language format to a respective embedding using the BAAI FlagEmbedding model to lower-dimensional vector of size 2014. We then use this embedding for the downstream prediction problem. 

\subsection{$\chi^2$-divergence and Correlation}
We consider two Gaussian distributions with the \emph{same} covariance and a mean shift: 
\(P=\mathcal N(\mu_P,\Sigma)\) and \(Q=\mathcal N(\mu_Q,\Sigma)\) with 
\(\mu_Q=\mu_P+\Delta\), \(\mu_P=\mathbf 0\), and
\(
\Sigma \;=\;
\begin{pmatrix}
\sigma_1^2 & c\,\sigma_1\sigma_2\\
c\,\sigma_1\sigma_2 & \sigma_2^2
\end{pmatrix},\qquad |c|<1.
\)
We fix \(\sigma_1=\sigma_2=1\) and draw a standardized mean shift \(z\sim \mathcal N(0,0.5^2 I_2)\) independently each trial, then set \(\Delta=z\odot(\sigma_1,\sigma_2)\).
For each correlation \(c\in\{-0.6,-0.3,0,0.3,0.6\}\) we run \(200\) trials.

\begin{table}[ht]
\centering
\small
\begin{tabular}{rrrrrr}
\hline
$c$ & mean & std & min & max \\
\hline
$-0.6$ & 5.288 & 29.283 & 0.005 & 379.781 \\
$-0.3$ & 1.286 & 3.919 & 0.004 & 50.341 \\
$0.0$ & 0.875 & 2.230 & 0.000 & 25.456 \\
$0.3$ & 1.229 & 2.394 & 0.001 & 22.107 \\
$0.6$ & 4.499 & 16.556 & 0.009 & 166.085 \\
\hline
\end{tabular}
\caption{$\chi^2$-divergence statistics over 200 trials with \(\sigma_1=\sigma_2=1\) and \(z\sim\mathcal N(0,0.5^2 I_2)\).}
\label{tab:chi2_equi_2d}
\end{table}

\noindent
In multimodal learning, stronger cross-modal correlation—whether positive or negative—magnifies how covariate shifts combine across modalities, leading to a larger chi-square divergence between source and target. When the modalities are uncorrelated, the divergence is smallest.

\end{document}